\RequirePackage[loading]{tracefnt}
\documentclass[sigconf]{acmart}

\usepackage{booktabs} 
\usepackage{amsmath,amssymb,amsfonts}
\usepackage{algorithm,tabularx}
\usepackage[noend]{algpseudocode}

\makeatletter
\newcommand{\multiline}[1]{%
	\begin{tabularx}{\dimexpr\linewidth-\ALG@thistlm}[t]{@{}X@{}}
		#1
	\end{tabularx}
}
\makeatother

\DeclareMathOperator*{\argmax}{arg\,max}

\copyrightyear{2020}
\acmYear{2020}
\setcopyright{acmlicensed}\acmConference[GECCO '20]{Genetic and Evolutionary Computation Conference}{July 8--12, 2020}{Cancun, Mexico}
\acmBooktitle{Genetic and Evolutionary Computation Conference (GECCO '20), July 8--12, 2020, Cancun, Mexico}
\acmPrice{15.00}
\acmDOI{10.1145/3377930.3390169}
\acmISBN{978-1-4503-7128-5/20/07}

\begin{document}
\title{Relatedness Measures to Aid the Transfer of Building Blocks among Multiple Tasks}

\author{Trung B. Nguyen}
\orcid{0000-0002-1990-8647}
\affiliation{
	\institution{Victoria University of Wellington, NZ}
	\streetaddress{PO Box 600, Wellington, New Zealand}
	\postcode{6140}
}
\email{trung.nguyen@ecs.vuw.ac.nz}

\author{Will N. Browne}
\orcid{0000-0001-8979-2224}
\affiliation{
	\institution{Victoria University of Wellington, NZ}
	\streetaddress{PO Box 600, Wellington, New Zealand}
	\postcode{6140}
}
\email{will.browne@vuw.ac.nz}

\author{Mengjie Zhang}
\orcid{0000-0003-4463-9538}
\affiliation{
	\institution{Victoria University of Wellington, NZ}
	\streetaddress{PO Box 600, Wellington, New Zealand}
	\postcode{6140}
}
\email{mengjie.zhang@ecs.vuw.ac.nz}

\renewcommand{\shortauthors}{Trung Nguyen et al.}

\begin{abstract}

Multitask Learning is a learning paradigm that deals with multiple different tasks in parallel and transfers knowledge among them. XOF, a Learning Classifier System using tree-based programs to encode building blocks (meta-features), constructs and collects features with rich discriminative information for classification tasks in an observed list. This paper seeks to facilitate the automation of feature transferring in between tasks by utilising the observed list. We hypothesise that the best discriminative features of a classification task carry its characteristics. Therefore, the relatedness between any two tasks can be estimated by comparing their most appropriate patterns. We propose a multiple-XOF system, called mXOF, that can dynamically adapt feature transfer among XOFs. This system utilises the observed list to estimate the task relatedness. This method enables the automation of transferring features. In terms of knowledge discovery, the resemblance estimation provides insightful relations among multiple data. We experimented mXOF on various scenarios, e.g. representative Hierarchical Boolean problems, classification of distinct classes in the UCI Zoo dataset, and unrelated tasks, to validate its abilities of automatic knowledge-transfer and estimating task relatedness. Results show that mXOF can estimate the relatedness reasonably between multiple tasks to aid the learning performance with the dynamic feature transferring.

\end{abstract}

%
%



\maketitle

\section{Introduction}
The ability to reuse knowledge among similar tasks allows humans to achieve skills and concepts through a few examples for each new problem. This motivates the advent of Multi-Task Learning (MTL), a learning paradigm where the learning system deals with multiple related tasks simultaneously with equal task priority \cite{caruana_multitask_1997,pan_survey_2010}. MTL aims to improve the learning performance of each task by transferring useful knowledge among related tasks.

However, existing MTL systems are generally restricted to related tasks where the contribution of common knowledge significantly dominates the adverse effect of the harmful signal coming from the unrelatedness of other tasks. Otherwise, the learning performances can become worse than those in separated traditional single-task learning. This limitation requires external knowledge on the tasks selected for an MTL system. On the contrary, the human ability of reusing knowledge is not bounded to related tasks because human intelligence can choose to relate a target task with appropriate ones. The hypothesis is that relatedness of any two problems estimated based on the overlap of their best-described patterns can be a guidance to transferring knowledge among tasks.

Evolutionary Computation learns optimisation tasks through building blocks that encourage transferring knowledge. Evolutionary MTL has been investigated with a variety of EC algorithms. Among them, the series of Multifactorial Evolutionary Algorithm \cite{gupta2015multifactorial,bali_multifactorial_2019} also offers the ability to prevent harmful interactions between distinct optimisation tasks using a matrix of random mating probability. However, MTL in optimisation tasks seeks transfer to guide search trajectories, while here we consider classification tasks where transfer of useful building blocks between systems is sought.

Learning Classifier Systems (LCSs) are an EC approach that inherently can divide-and-conquer to split a complex problem into simpler sub-problems, i.e. niches. The learning capability of LCSs can be scaled up with the use of Genetic Programming-like tree-based programs, called Code Fragments (CFs) used in the conditions (and/or actions) of the classifiers \cite{iqbal_reusing_2014}. CFs are a representation of features that can facilitate feature transfer and thereby MTL. CFs or CF-based functions are the media of transferring knowledge in modern LCSs \cite{iqbal_reusing_2014,alvarez_human-inspired_2016} to solve large-scale and complex Boolean problems. A recent work on using CFs in LCS's rule conditions introduced the concept of the Observed List (OL) \cite{8789950,10.1145/3321707.3321751}. This list includes the most applicable CFs, which contain the most discriminative information for the target task. Incorporating the OL in an accuracy-based LCS that utilises CFs created the XOF system. It is hypothesised that leveraging OLs to estimate the task relatedness can assist sharing knowledge among tasks with XOF.

In this paper, we propose a system of multiple XOFs, called mXOF, that can solve different problems simultaneously and automatically detect the common characteristics of the problems to facilitate transferring features among XOFs. Each XOF in mXOF learns one problem. The comparison of two OLs is used to estimate the relatedness of two problems. The objectives of this research are:
\begin{enumerate}
	\item To create an MTL system that automatically shares features among related tasks by estimating the asymmetric relatedness of tasks. The probability $p_{i,j}$ of transferring features from a problem $i$ to a problem $j$ is determined by the estimated relatedness, where $p_{i,j}$ and $p_{j,i}$ are not identical.
	\item To facilitate estimating the relatedness between tasks by using the similarity of their OLs.
	\item To investigate the ability of mXOF to handle arbitrary multiple tasks, including related and unrelated tasks. This can further validate the automation of mXOF in sharing features among tasks. If the system can estimate a low relatedness between two unrelated problems, the learning signals of these two problems should not interfere with the learning processes of each other. Handling arbitrary tasks contributes to general AI systems with continual learning \cite{thrun1995lifelong,hassabis2017neuroscience}.
\end{enumerate}

The specific context of this work is that a learner (robot or computer), is learning to recognise different objects (classes) in parallel using signals from the same sensor as the input data. In the very beginning, recognising all objects starts with the original data from the sensor, named as $(D_0,D_1,D_2,...)$. At this stage, there is no divergence between the tasks except for the expected output ($1/0$) because the learner sees all objects as the raw input signal. The progressive learning grows high-level building blocks from the original input by feature construction inherent in CFs, which cause the recognition tasks to diverge. The mission of mXOF is to track the relatedness among the recognition tasks to automate the transfer of the grown building blocks to improve the learning process of each task and link related tasks together.

The system will be tested on multiple scenarios to validate its capability to automate sharing features among tasks. First, multiple hierarchical problems \cite{butz2006rule}, which share the equivalent base-level patterns, will validate the ability to transfer CFs among highly related tasks. Also, a scenario of two relatively unrelated tasks, i.e. 11-bit Even Parity problem and 10-bit Carry problem, will test the ability of mXOF to transfer features among unrelated tasks selectively. Finally, a practical multi-class classification problem will be used to evaluate mXOF as a multi-class classifier. Because non-transfer learning benchmark datasets have been constructed for independent tasks, so are often unrelated, the UCI Zoo dataset has been repurposed by converting to multiple binary classifications to test whether mXOF can discover possible relatedness among seven classes of the dataset. This also experiments mXOF without advanced knowledge on the relatedness among classes. 


\section{Background}\label{sect:backgr}

\subsection{Learning Classifier Systems}
LCSs refer to a family of algorithms following a concept formalised by Holland's work on adaptation \cite{holland_adaptation*_1976}. Generally, an LCS interacts with an environment to update and evolve a rule-based population through evolutionary methods \cite{urbanowicz_introduction_2017}. XCS is a reinforcement-learning implementation of LCS that uses accuracy-based fitness to evolve an online-learning agent with a population of cooperative classifiers \cite{butz_algorithmic_2000,wilson_classifier_1995}. The workflow of XCS, when receiving an environment state, includes the following main steps: (1) matching to form a match set of classifiers applicable to the state; (2) classifier initialisation in covering if no classifier matches; (3) action selection from the match set to form an action set; (4) rule discovery to generate classifiers using Genetic Algorithm-adapted operators \cite{holland1992genetic}; and (5) classifier parameter update following the environmental response to selected actions. XCS has a unique divide-and-conquer property as the reproduction process and classifier update occur within local niches, i.e. the action sets $[A]s$. This property results in XCS's capability to divide a complex problem into sub-problems (i.e. niches) that can be solved more easily. MTL can benefit from this property as solved niches in a task can also provide useful patterns for other tasks.


A rule in XCS population is in the form of ``if \textit{condition} then \textit{action}''. In classification tasks, the rule action corresponds to the predicted class. Traditionally, XCS uses the ternary representation $\{0,1,\#\}$ for rule conditions. This representation limits the complexity of patterns that rule conditions can produce. Thus, it inhibits XCS from describing complex patterns in hierarchical problems. Fortunately, XCS allows encoding its rules using rich representations, such as tree-based programs \cite{ahluwalia1999genetic,pier1999extending,10.1007/978-3-540-88138-4_3}.

\subsection{CF-based XCSs}\label{ssect:CFXCSs}

Code Fragments (CFs) are a form of binary tree-based programs that were initially used in Boolean problems with a depth limit of $2$ \cite{iqbal_reusing_2014}. CFs are comprised of internal nodes and terminal (leaf) nodes. An internal node is a function from a function set, while a leaf node corresponds to an original feature, i.e. an attribute from environment states, or a previously learned CF. In Boolean domains, the function set usually contains general binary operators $\{AND, OR, NOT, NAND, XOR\}$.

Using CFs allows XCS to produce complex patterns which can result in more straightforward decision boundary in classification problems. With the aid of transfer learning \cite{pan_survey_2010} and layered learning \cite{stone_layered_2000}, XCS can scale to solve larger-scale problems that used to be intractable for standard XCS. Iqbal et al. introduced the reusabilities of learned CFs in leaf nodes of XCSCFC to scale the learned knowledge in rule conditions \cite{iqbal_reusing_2014}. XCSCFC successfully solved 135-bit Multiplexer by reusing learned CFs from smaller-scale Multiplexer problems. Alvarez et. al extended the reusability of CFs to use rule populations of solved problems as rule-set functions for use in internal nodes \cite{10.1145/2598394.2611383}. With layered learning, XCSCF* \cite{alvarez_human-inspired_2016} can obtain a general logic of the Multiplexer domain by decomposing it into subtasks and combining learned knowledge from those subtasks when considering the 6-bit Multiplexer problem as training data.

\subsection{XOF and the New Online-Feature generation module}

The Online-Feature generation (OF) is an extension to XCS using CF-conditions, where such a system is termed XOF \cite{8789950,10.1145/3321707.3321751}. It can reliably grow complex tree-based features (CFs) from the original data features. This method is similar to growing Genetic Programming trees \cite{koza_genetic_1992}. XOF maintains a CF population with a preferable list, called the Observed List (OL). The OL contains the CFs with the highest discriminative information of the target problem. The idea behind XOF is to grow tree-based features from the most useful CFs in the Observed List (OL) to construct useful high-level CFs for rule conditions in tree-based XCS. This is expected to find more complex patterns that can simplify the task by constructing more accurate and general classifiers (high-fitness classifiers). The ability of CFs to create more accurate and generalised classifiers defines CF-fitness ($cf.f$) as the applicability of CFs:

\begin{equation}
	cf.f=\frac{cl.f}{ \text{number of CFs in }cl},
\end{equation}
where $cf$ is the CF with CF-fitness to be evaluated, classifier $cl=\argmax_{cl \mid cf \in cl}cl.f$, and $cl.f$ is the fitness of classifier $cl$. This CF-fitness rates a CF according to the accuracy and generality per CF of the highest-fitness classifier containing the CF without caring about the complexity of the CF. 

\begin{figure}
	\vspace{-3mm}
	\centering
	\includegraphics[width=0.9\columnwidth]{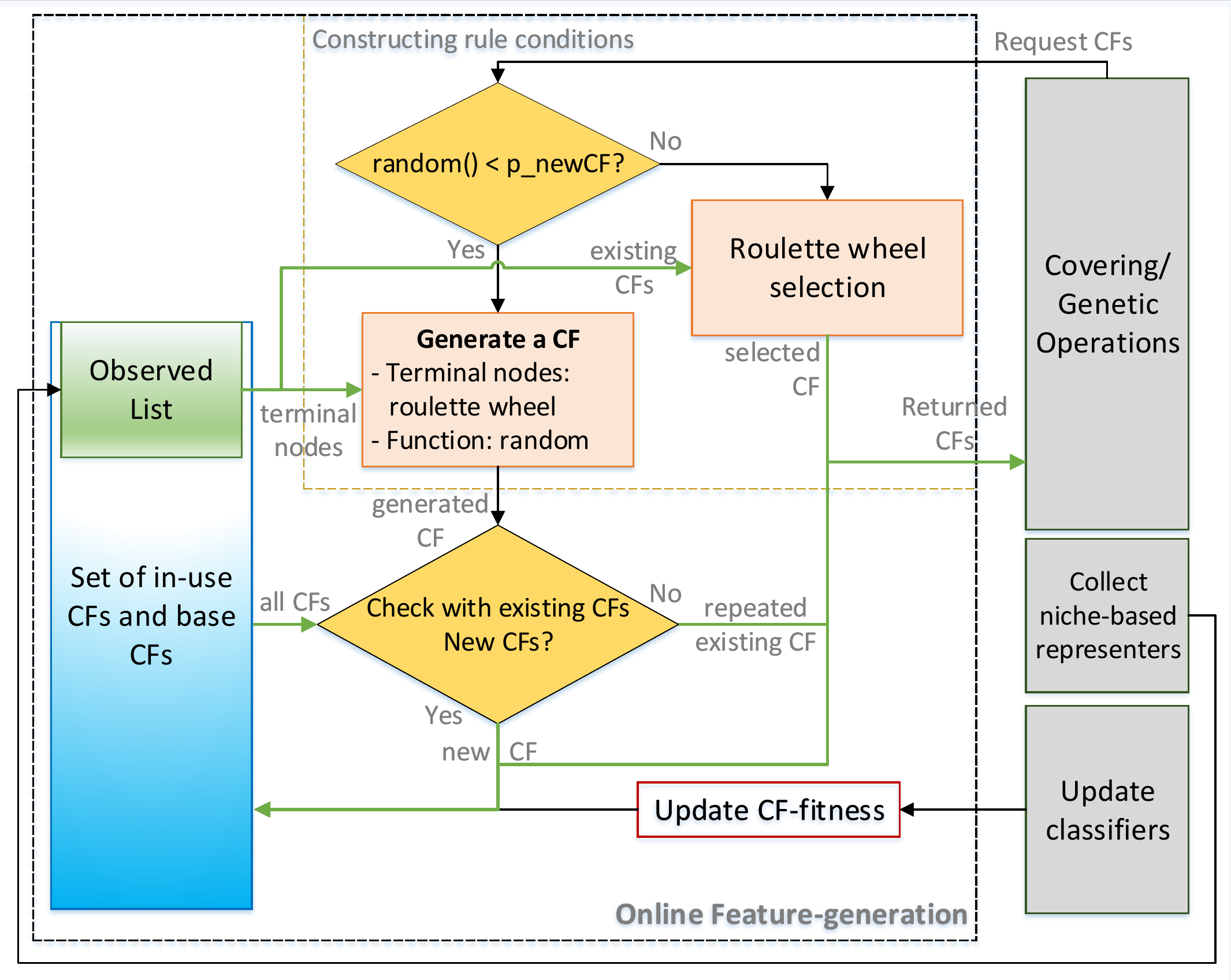}
	\caption{The OF module. It returns a CF when being requested by general processes of XCS. The returned CF can be an existing one selected by Roulette Wheel or a newly generated one. In both cases, the OF relies on the OL.}
	\label{fig:OF}
	\vspace{-3mm}
\end{figure}

A recent upgrade of the OF module (see Figure \ref{fig:OF}) simplified the OL update by collecting the niche-based highest fitness-rate classifiers \cite{nguyen2020complexity}. This was achieved by introducing the ``rule-fitness rate'' for CFs, which is equivalent to the fitness per complexity of classifiers: 

\begin{equation}
	cl.f\_rate=\frac{cl.f}{cl.complexity}.
\end{equation}
The complexity of a rule is the total number of leaf nodes in all CFs in the condition. The fitness per complexity additionally punishes the ability of CFs to create high-fitness classifiers by the classifier complexity. Accordingly, the new OF module collects CFs for the OL from the classifiers with the highest fitness per complexity in the action set of XCS. Also, the CF-fitness updates of a CF follow the fitness per complexity of the classifier with the highest fitness per complexity that contains the CF:

\begin{equation}
	cf.f=\max_{cl \mid cf \in cl} \frac{cl.f}{cl.complexity},
\end{equation}
This CF-fitness represents the highest fitness per leaf node that the CF can produce among generated rules, called rule-fitness rate. In short, the OF module rates generated CFs based on their efficiency of using binary operators to combine the input attributes to produce accurate and general classifiers. The rule-fitness rate allows XOF to build more complexity-efficient patterns.

\section{Multi-task Learning with \lowercase{m}XOF}\label{sect:mXOF}

\begin{figure}
	\centering
	\includegraphics[width=0.80\columnwidth]{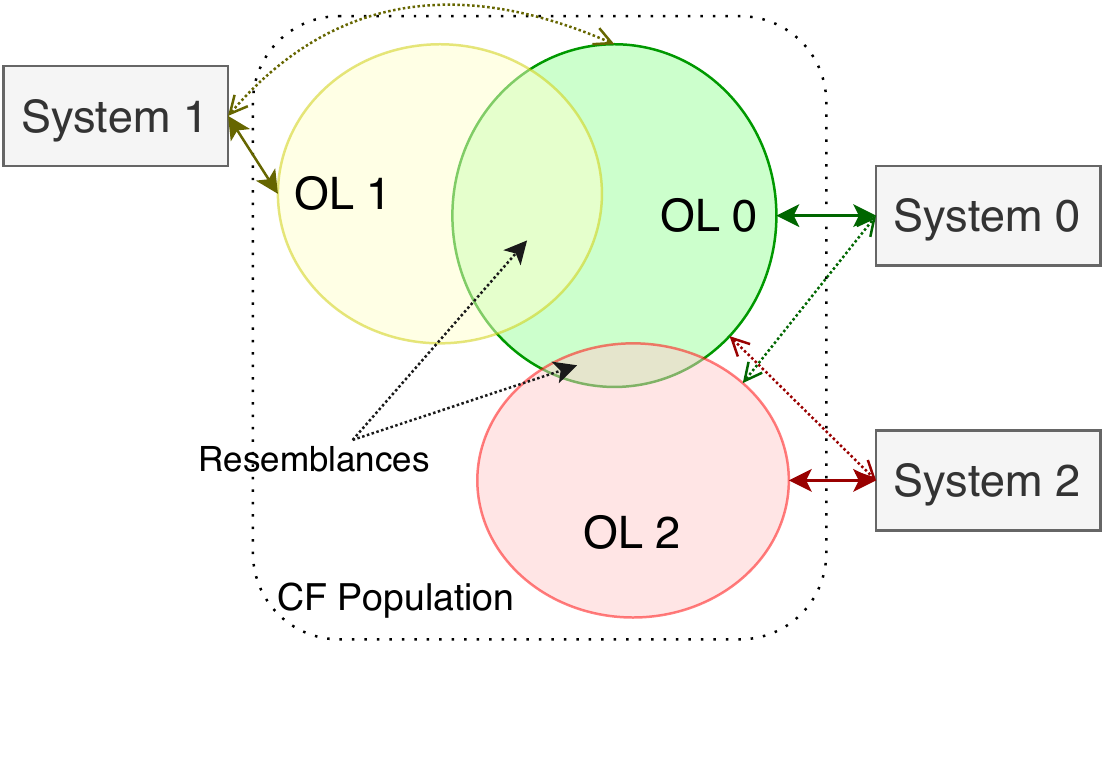}
	\caption{A mXOF with three tasks. A large resemblance between two OLs indicates a high relatedness between two tasks, which is the guidance of automatic CF sharing.}
	\label{fig:mXOF}
	\vspace{-3mm}
\end{figure}

We propose a mXOF system where each XOF learns one of the separated tasks together with the other XOFs, i.e. one XOF per task. Figure \ref{fig:mXOF} illustrates a case of mXOF with three systems and three tasks. The sharing of CFs and the relatedness measurement among systems mutually support each other during the learning processes of multiple systems. In this paper, we will use the same identification for a system and its corresponding task because each system works on one task.

All systems utilise a common CF population. The CF population here serves similar purposes as the CF population does in a single-XOF system. The population enables tracking generated CFs and updating their CF-fitness in all systems. Furthermore, it also links equivalent CFs among single-XOF systems, which enables the estimation of task relatedness and transferring CFs among systems.

\subsection{Asymmetric Fitness-weighted Relatedness}

The commonality of CFs between the OLs of two tasks is hypothesised to indicate the relatedness between the tasks. Specifically, CFs in the OL of a system are the most discriminative patterns among generated features, which are useful in constructing accurate and generalised rules, i.e. high-fitness rules. Therefore, two tasks with common discriminative patterns in the intersection of two OLs are considered related tasks.

Conversely, the relatedness of a system $a$ to another system $b$ is to estimate the possible applicability of CFs produced in system $a$ to system $b$ (see Section \ref{ssect:CFXCSs} for the applicability of CFs). The fitness-weighted relatedness of system $a$ to system $b$ corresponds to the CF-fitness portion of sharing CFs between $a$ and $b$ in the total CF-fitness of CFs in the OL of system $a$:

\begin{equation}
	RelSS(a,b) = \frac{\sum_{cf_i \in OL_a \land OL_b} cf_i.f(a)}{\sum_{cf_j \in OL_a}cf_j.f(a)},
\end{equation}
where $OL_{a}$ and $OL_{b}$ are two OLs of system $a$ and $b$ respectively, and $cf.f(a)$ is the CF-fitness of $cf$ in system $a$. This asymmetric definition of task relatedness is a statistical estimation of the applicability of any CF from the OL of system $a$ to system $b$. The maximum relatedness is $1$ when all CFs in the OL of the source task are applicable to the target task, and is $0$ when no CFs in the OL of the source task is relevant to the target task. The asymmetric property is desirable because the applicability of features between two tasks is generally not symmetric. For example, when a task can be a subset of another task, almost all the features from one task is applicable to the other task while the applicability in the opposite direction can be minimal.

\subsection{Automatic Transferring of CFs}
The CF transferring among tasks is automatically driven by their relatedness. This mechanism benefits the learning processes of involved systems in multiple ways. First, there is no need for human intervention as in transfer learning or layered learning, such as selecting a sequence of highly related tasks and criteria of features for transferring. Second, this enables the dynamic probability of transferring features among tasks. This property is desirable because, when the learning process of XOF generates CFs, the applicability of CFs from one task to another and the relatedness of tasks usually change dynamically. Lastly, feature transferring provide individual systems during evolving with external knowledge that can help escape local optima.

In mXOF, a system uses transferred CFs as well as existing CFs in its OL when selecting CFs for covering new rules or mutating a rule condition. Algorithm \ref{alg:cf_transfer} illustrates how the selection procedure works. Whenever a target system $a$ queries an existing CF, it will select from its OL plus at most one external CF from other systems. The external CF is selected from a set $S_{ecf}$ of all CFs in the OLs of other systems that satisfy two relatedness criteria. The first criterion is that the source system providing external CFs must be more related to the target system $a$ than a threshold $r\_thres$. This threshold is drawn from a uniform distribution (step $2$, discussed later). Second, these CFs themselves have to be potentially applicable to the target task $a$. We introduce a statistical expectation of its relatedness to the target task. The expected relatedness of an external CF $cf_j$ is actually the relatedness of the source task containing $cf_j$ adjusted by the rate between CF-fitness of $cf_j$ and the average CF-fitness of the common (shared) CFs (between the source task $i$ and the target task $a$) on the source task $i$:

\begin{align*}
	RelCfS(cf_j,a)=\frac{cf_j.f(i)}{\sum_{cf \in C(i,a)} cf.f(i)/len(C(i,a)}*RelSS(i,a),
\end{align*}
where $C(i,a)=OL_i \land OL_a$. This check can also be interpreted that the performance of the candidate $cf_j$ should be comparable with the performances (on the source task) of shared CFs between two tasks. If the relatedness of the external CF satisfies the threshold $r\_thres$, $S_{ecf}$ will append this CF with its adjusted vote (applicability) shown in Step $10$. Finally, a Roulette Wheel selection chooses a CF from $S_{ecf}$ using adjusted votes of external CFs. 

\begin{algorithm}
	\vspace{-1mm}
	\begin{algorithmic}[1]
		\State Collection of external CFs from other systems $S_{ecf}=\emptyset$
		\State Draw a relatedness threshold from a uniform distribution $r\_thres=max(0.1,uniform(0,1))$
		\For{system $i \neq a$}
		\If{$RelSS(i,a)>=r\_thres$}
		\State Common CFs in the $OL_i$ and $OL_a$: $C(i,a)=OL_i \land OL_a$
		\For{$cf_j \in OL_i$, the OL of $i$}
		\If{$cf_j \notin OL_a$}
		\State \multiline{Relatedness of $cf_j$ to system $a$: \\$RelCfS(cf_j,a)=RelSS(i,a)*(cf_j.f(i)/$\\$(\sum_{cf \in OL_i \land OL_a} cf.f(a)/len(C(i,a))))$}
		\If{$RelCfS(cf_j,a)>=r\_thres$}
		\State \multiline{Compute adjusted vote of $cf_j$: $vote_{adj}(cf_j,a)=cf_j.f(i)*$\\$(\frac{\sum_{cf \in (OL_i \land OL_a} cf.f(a)}{\sum_{cf \in (OL_i \land OL_a} cf.f(i)}*RelSS(i,a))$}
		\State \multiline{Add $cf_j$ to external selections $S_{ecf}.add(cf_j)$ with its adjusted vote}
		\EndIf
		\EndIf
		\EndFor
		\EndIf
		\EndFor
		\State Select an external CF from $S_{ecf}$ using roulette wheel selection and their adjusted votes $cf_e=RW(S_{ecf})$
	\end{algorithmic}
	\caption{Transferring a CF $cf_e$ from other systems to reuse in target system $a$. $cf_e$ with its adjusted vote $vote_{adj}(cf_e,a)$ will be another candidate for system $a$ when selecting existing CFs to construct rules.}
	\label{alg:cf_transfer}
	\vspace{-1mm}
\end{algorithm}

The relatedness threshold $r\_thres$ drawn from a uniform distribution (step $2$) represents the stochastic selectivity of the learner in sharing CFs among tasks. With this philosophy, we use this threshold as a common filter for both tasks and CFs. This also simplifies parameter setting, where future work could consider the effects of separate tuned thresholds. Using the uniform distribution for the sharing selectivity might not filter out all negative transfer. However, when there are more and more tasks, the most related tasks would always stand more probability to share with one another.

The vote (applicability) of an external CF is adjusted according to the performance of the shared CFs between the two OLs on the source task and the target task, which is the rate between the total CF-fitness of shared CFs in the target task and the corresponding amount in the source task. The better performance of shared CFs on the target task, the higher adjusted vote the external CF can get. This adjusted applicability is also used for the external CF when competing with existing CFs in the OL of the target task for constructing rules.

\subsection{mXOF for Multi-class/Multi-label Classification}

Multi-class and multi-label classification can be converted into multiple binary classification problems, where each class/label corresponds to a binary classifier, i.e. an XOF. All binary classifiers work synchronously on each same instance with converted ground truth. That is, while each system receives the same environment state (instance input) each iteration, only one that works on the true class of the instance should output $1$ to receive maximum reward $1000$ and other systems should output $0$ for the maximum reward. mXOF becomes multiple recognition systems, where each XOF recognises one class/label.

When evaluating an instance on the test set, each binary classifier produces a probability, or a confidence level, that this instance belongs the class corresponding to the classifier. The probability is necessary in case multiple binary classifiers output $True$ on an instance in multi-class classification tasks. We implemented this probability based on the prediction array of XOF. The probability of action $1$, which means the corresponding class is detected, is:

\begin{equation}
	P(1)=\frac{\sum_{cl \in [A] | cl.action=1} cl.prediction*cl.f}{\sum_{cl \in [A]} cl.prediction*cl.f}.
\end{equation}
This is the rate of the total fitness-weighted prediction of action $1$. Thus, the probability of action $0$ is $P(0)=1-P(1)$. When more than one binary classifiers produce the same highest probability, mXOF randomly selects a class from the classes of such binary classifiers. In the case of multi-label classification, each system in XOF can choose the values of exploited actions (actions with highest predicted payoff in the prediction array) for its target label.

The possible downside of converting multi-class classification into multiple binary classifications is that one balanced dataset can become multiple imbalanced data. However, we expect that this problem does not have much influence on the result because XCS can manage to balance its niches very well \cite{urbanowicz_introduction_2017}. On the contrary, using mXOF for multi-class and multi-label classification provide relatedness among classes/labels. The class/label relatedness provides an insight knowledge on the relationships among target classes/labels. 


%


\section{Experimental Results}\label{sect:exp_res}

In this section, we compare mXOF that simultaneously addresses several problems with individual XOF on the same but separated problems. There are two criteria for comparisons: the learning performance as well as the discovery of complexity-efficient CFs. All experiments were evaluated based on the average of $30$ independent runs.

All parameters of each system in mXOF are equal to corresponding ones in single XOF for the same problems. Except for the rule population size and stopping iteration, we used a general configuration of XOF \cite{10.1145/3321707.3321751} for other parameters in these experiments: the learning rate $\beta=0.2$; the crossover rate is $\chi=0.2$; the mutation rate is $\mu=0.9$; the experience thresholds for deletion is $\theta_{del}=20$; the initial fitness of covered classifiers are $F_{init}=0.01$; the probability of specificness $p_{spec}=0.25$ with maximum rule-condition length set at twice the number of original input attributes; and the experience thresholds for subsumption $\theta_{sub}=50$. The learning performances in these experiments counted exploration trials/instances. 

The implementation is in multi-threaded Python with a progress synchroniser to assure all systems in mXOF experience the same number of iterations during their learning processes. The purpose of this progress synchronisation is to produce a more stable and reliable evaluation.

To track and evaluate approximately the generality rate of the most complexity-efficient CFs in an XOF, we collect the most efficient classifiers in action sets, i.e. highest fitness per complexity, that satisfy: 

\begin{equation}
	cl.f/cl.complexity>=0.8*\max_{cl \in [A]}cl.f/cl.complexity.
\end{equation}
These are also the classifiers selected for collecting the OL. The tracked generality rate is the average generality rate of the most efficient and accurate classifiers with $cl.error=0$. Generality rates of classifiers are estimated as follows:

\begin{align*}
&cl.generality=\frac{cl.matches}{cl.matches+cl.no\_matches}, \text{ thus}\\
&cl.generality\_rate=\frac{cl.generality}{cl.complexity},
\end{align*}
where $cl.matches$ and $cl.no\_matches$ respectively track the numbers of times classifier $cl$ matches and does not match all instances since it was created, and therefore the part $cl.matches/(cl.matches+cl.no\_matches)$ provides the generality of classifier $cl$ as it tracks the probability that classifier $cl$ matches any instance.

\subsection{MTL with Hierarchical Boolean Problems}

Hierarchical Boolean problems are problems that combine a Boolean problem at the top-level with successive 3-bit Even-parity problems at the bottom-level \cite{butz2006rule}. Because of this combination, hierarchical problems have highly complex underlying patterns. Even though these problems can be small in scale, the search spaces in solving them are much higher than other Boolean problems, such as Multiplexer, at the same scale because they require complex combination of attributes. Discovery of these patterns can simplify the search of decision boundaries, i.e. the rules of XOF. These experiments can be considered to involve related problems because at some stage of learning, XOF can construct CFs covering the bottom-level 3-bit Even-parity problem, which are common among problems.

\begin{figure}
	\vspace{-3mm}
	\centering
	\includegraphics[width=0.9\columnwidth]{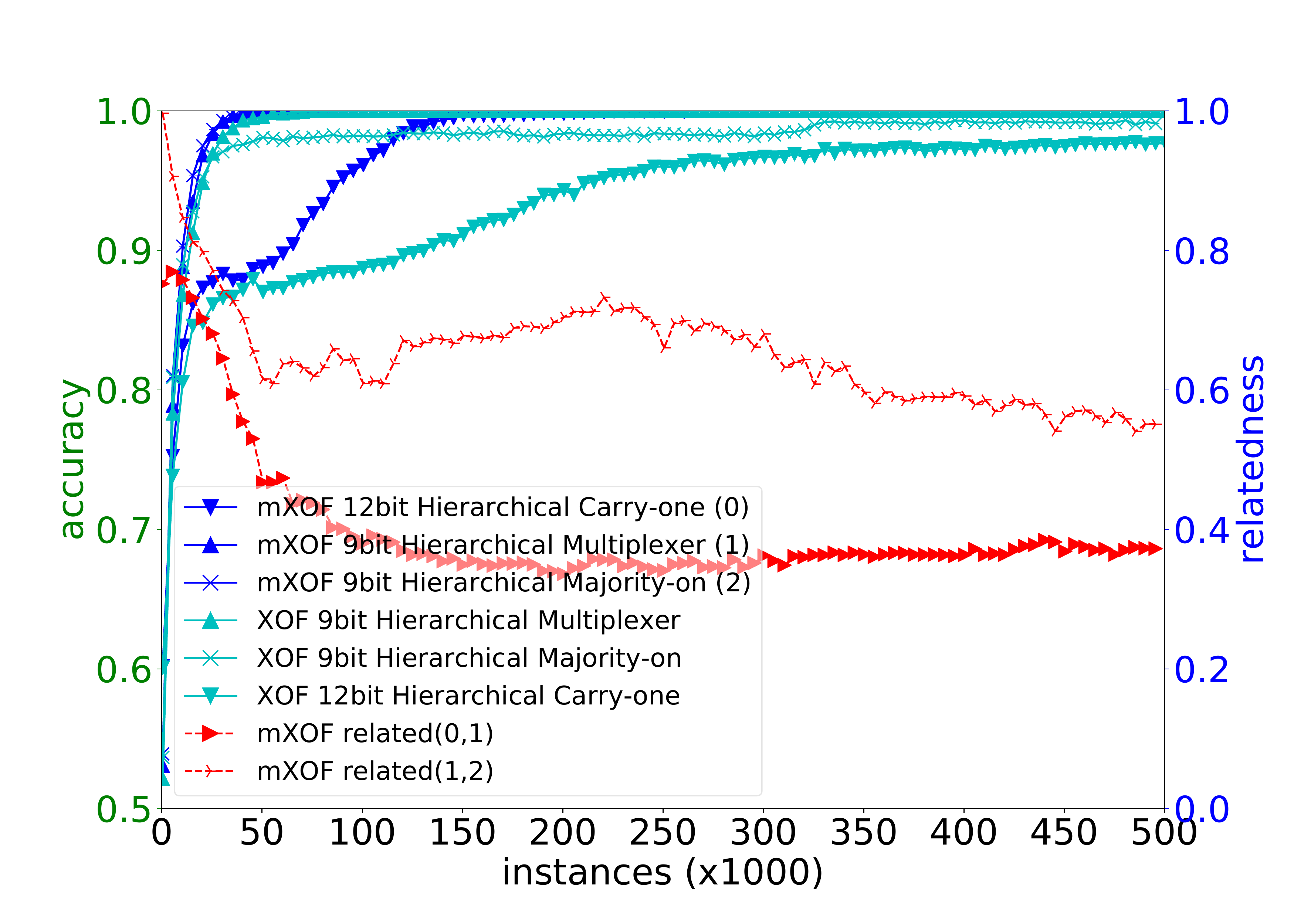}
	\caption{Learning performance on multiple small-scale hierarchical problems. The red lines records the relatedness among tasks. Note: $0$ is Hierarchical Carry-one, $1$ is Hierarchical Multiplexer, and $1$ is Hierarchical Multiplexer.}
	\label{fig:hiers9}
	\vspace{-3mm}
\end{figure}

\begin{figure}
	\vspace{-2mm}
	\centering
	\includegraphics[width=0.9\columnwidth]{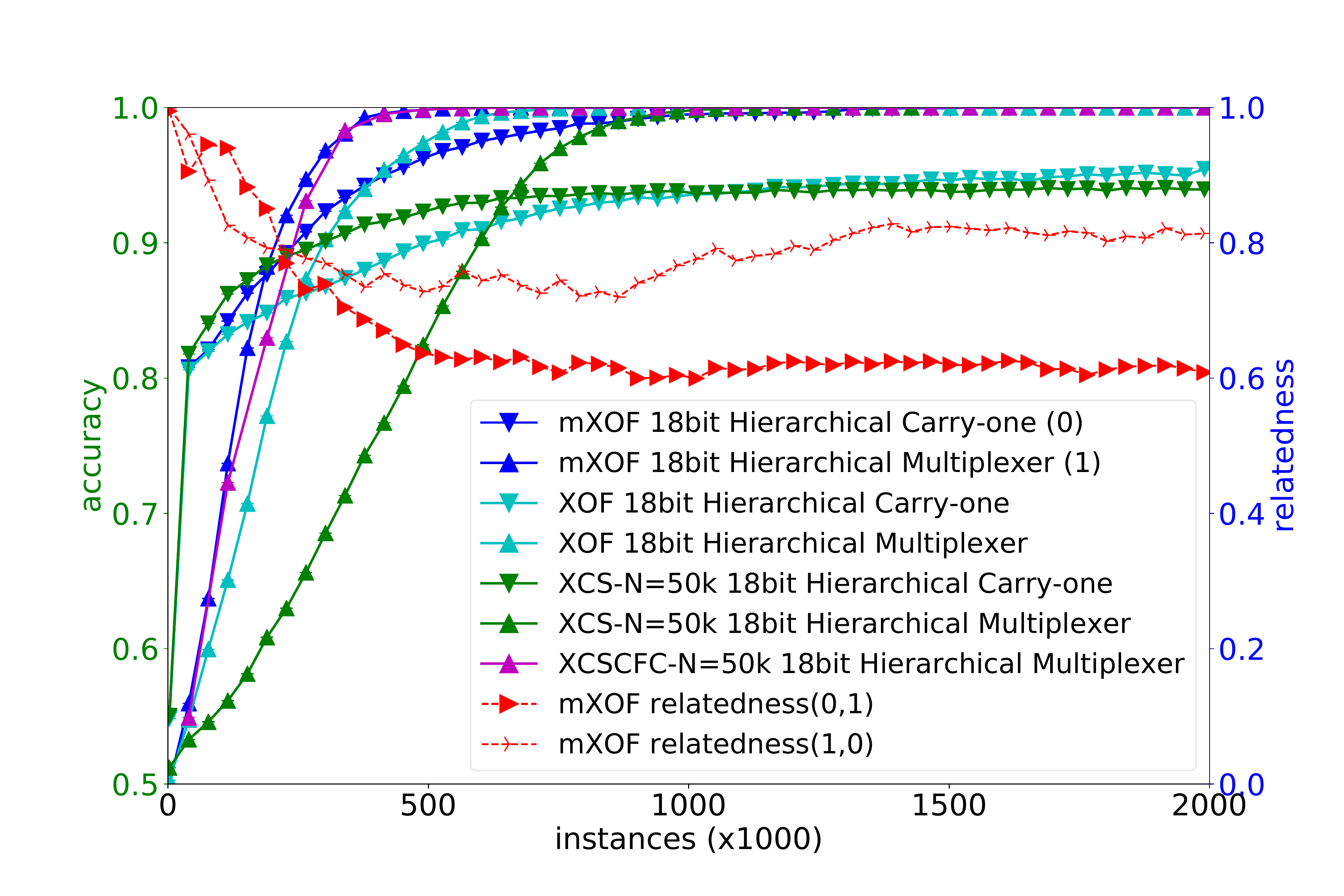}
	\caption{Learning performance on multiple 18-bit hierarchical problems. The red lines records the relatedness between two tasks. Note: $0$ is Hierarchical Carry-one, $1$ is Hierarchical Multiplexer.}
	\label{fig:hiers18}
	\vspace{-3mm}
\end{figure}

We first evaluated mXOF on two set of multiple hierarchical problems. The first set includes 12-bit Hierarchical Carry-one, 9-bit Hierarchical Multiplexer, and 9-bit Hierarchical Majority-on problems. All individual systems of mXOFs and single XOFs used the same population size of $N=2000$ (Figure \ref{fig:hiers9}). The second experiment has two larger-scale problems including 18-bit Hierarchical Carry-one and 18-bit Hierarchical Multiplexer problems (Figure \ref{fig:hiers18}). In this experiment, individual systems of mXOF and single-system XOFs all had the same population size of $N=20000$. We also compared mXOF with XCSCFC and XCS with the same higher population size of $N=50000$. These problems require constructing relevant and complex patterns to simplify decision boundaries. In both experiments, the performances of mXOF for each of these problems are superior to those of single-system XOFs on corresponding problems. The MTL system also achieves higher accuracies compared with the single-system method except for XCSCFC. This system has a similar performance with mXOF on the 18-bit Hierarchical Multiplexer problem, but it has a higher population size and also requires transfer learning \cite{iqbal_reusing_2014}.

The average relatedness of 18-bit Hierarchical Carry-one problem to 18-bit Hierarchical Multiplexer problem is also illustrated on Figure \ref{fig:hiers18}. It starts at high values and declines quickly when the accuracies of these problems progresses quickly to $100\%$. After reaching the maximum accuracy, the average relatedness maintains at around $0.6$. 

\begin{figure}
	\vspace{-3mm}
	\centering
	\includegraphics[width=0.80\columnwidth]{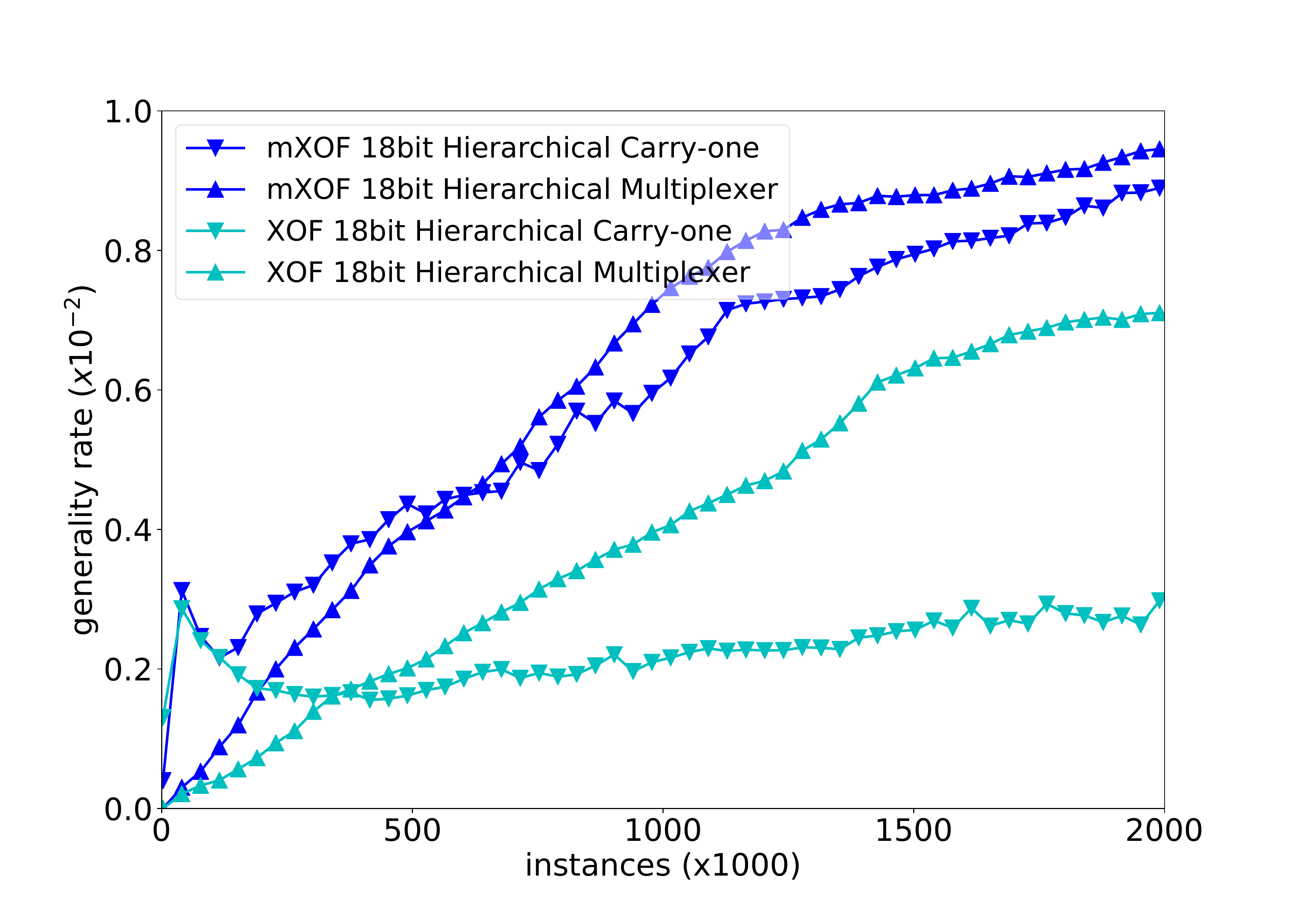}
	\caption{Generality rates on 18-bit hierarchical problems.} 
	\label{fig:hiers18_genex}
	\vspace{-3mm}
\end{figure}

\begin{figure}
	\vspace{-1mm}
	\centering
	\includegraphics[width=0.90\columnwidth]{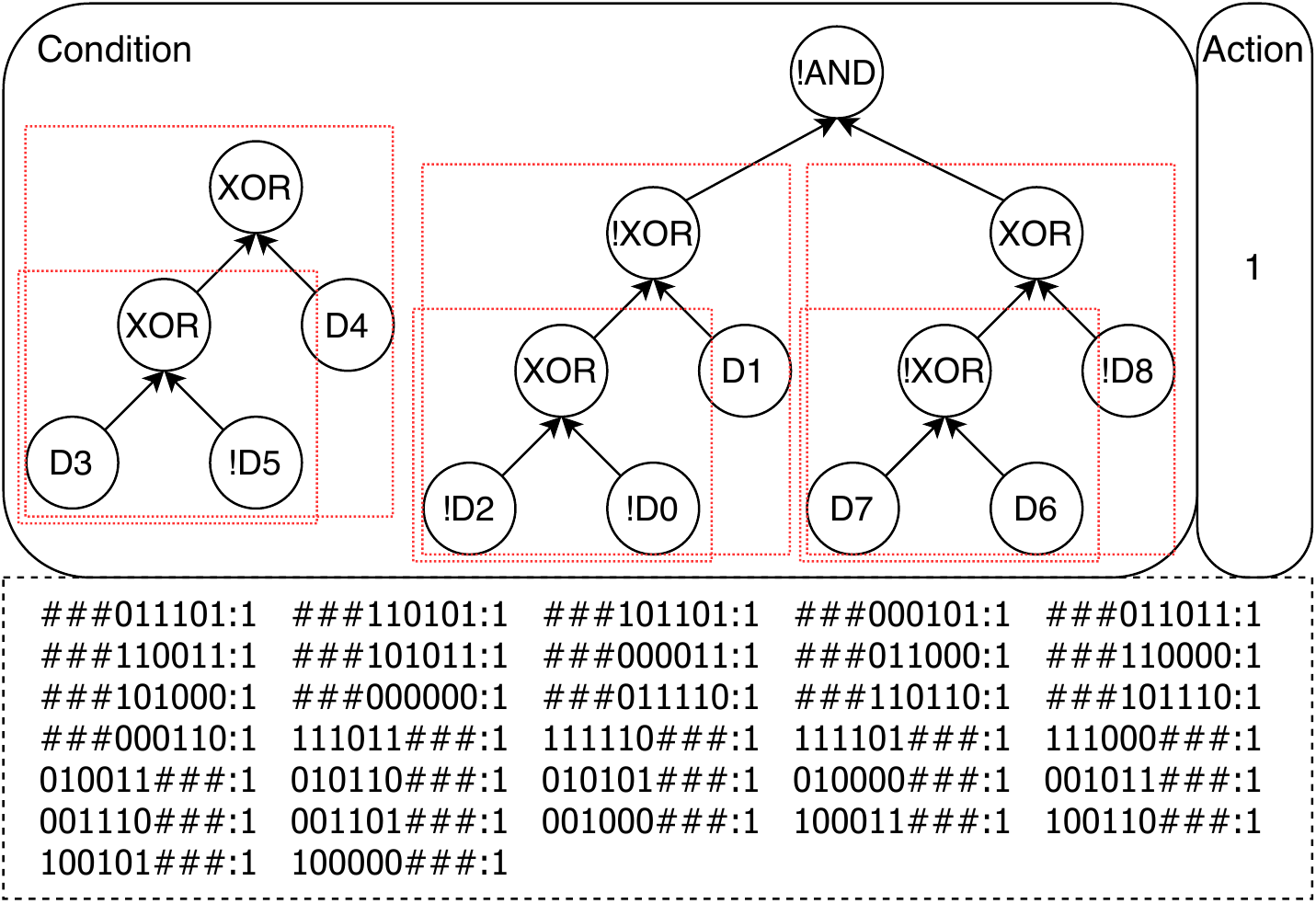}
	\caption{An optimal rule (prediction $1000$) of the 9-bit Hierarchical Multiplexer problem with its equivalent ternary rules. With the generality of $0.375$ and the complexity of $9$, its generality rate is $1/24$. The CFs in red boxes are reusable grown-patterns for any Hierarchical Boolean problem.}
	\label{fig:CF_exp}
	\vspace{-3mm}
\end{figure}

The progress of discovering complexity-efficient CFs for both mXOF and XOF is shown in Figure \ref{fig:hiers18_genex}. The generality rate of exploit classifiers evolves faster in mXOF since the discovery of complex and efficient CFs in all systems supports each other.
Specifically, the optimal set of complexity-efficient CFs required to solve the 9-bit Hierarchical Multiplexer problem must include at least three CFs covering the three non-overlapped 3-successive-bit chunks, which correspond to three underlying bottom-level 3-bit Even-parity problems, or the combinations of these CFs. Such optimal CFs that can cover a 3-bit Even-parity problem must combine 3-bit input using $XOR$ operator (with an arbitrary amount of the $NOT$ function). These CFs are the reusable grown-patterns among all hierarchical Boolean problems. The sample rule in Figure \ref{fig:CF_exp} is optimal in terms of the complexity efficiency for the 9-bit Hierarchical Multiplexer problem. This rule contains various constructed patterns (in red dashed boxes) that are transferable across the hierarchical problems. These patterns include the CFs covering the bottom-level problems and the CFs constructed in the middle way (in three smaller red boxes).

These shared patterns enable rules using them to generalise and so cover a larger set of instances. The sample rule in Figure \ref{fig:CF_exp} is equivalent to $32$ ternary rules. Because the niche of each ternary rule cross partly with another, the coverage of these $32$ rules is equivalent to $24$ ternary rules. Therefore, the sample rule in Figure \ref{fig:CF_exp} covers $24/2^6=3/8$ of the instance space, where $6$ is the number of specified bits in these rules. The practical value of the optimal generality rate of 9-bit Hierarchical Multiplexer problem is $3/8*1/9=1/24\approx0.0416$.

\subsection{MTL with Low Relatedness}\label{ssect:low_related}

In this experiment, we evaluated mXOF in its ability to prevent negative interactions among unrelated or slightly related tasks. In this experiment, we chose two sets of Boolean problems: (1) 37-bit Multiplexer and 11-bit Even-parity problems; and (2) 10-bit Carry-one and 11-bit Even-parity problems. While 37-bit Multiplexer problem does not require complex patterns, 10-bit Carry-one problem and especially 11-bit Even-parity problem can benefit from constructing complex CFs \cite{8789950}. In the second experiment, some hierarchical patterns from 10-bit Carry-one problem can be useful for 11-bit Even-parity problem, e.g. $D_0\times(!D_5)$ and $D_7\times(!D_2)$. However, most of the hierarchical patterns from 11-bit Even-parity problem contain no discriminative information for 10-bit Carry-on problem.

\begin{figure}
	\vspace{-3mm}
	\centering
	\includegraphics[width=0.9\columnwidth]{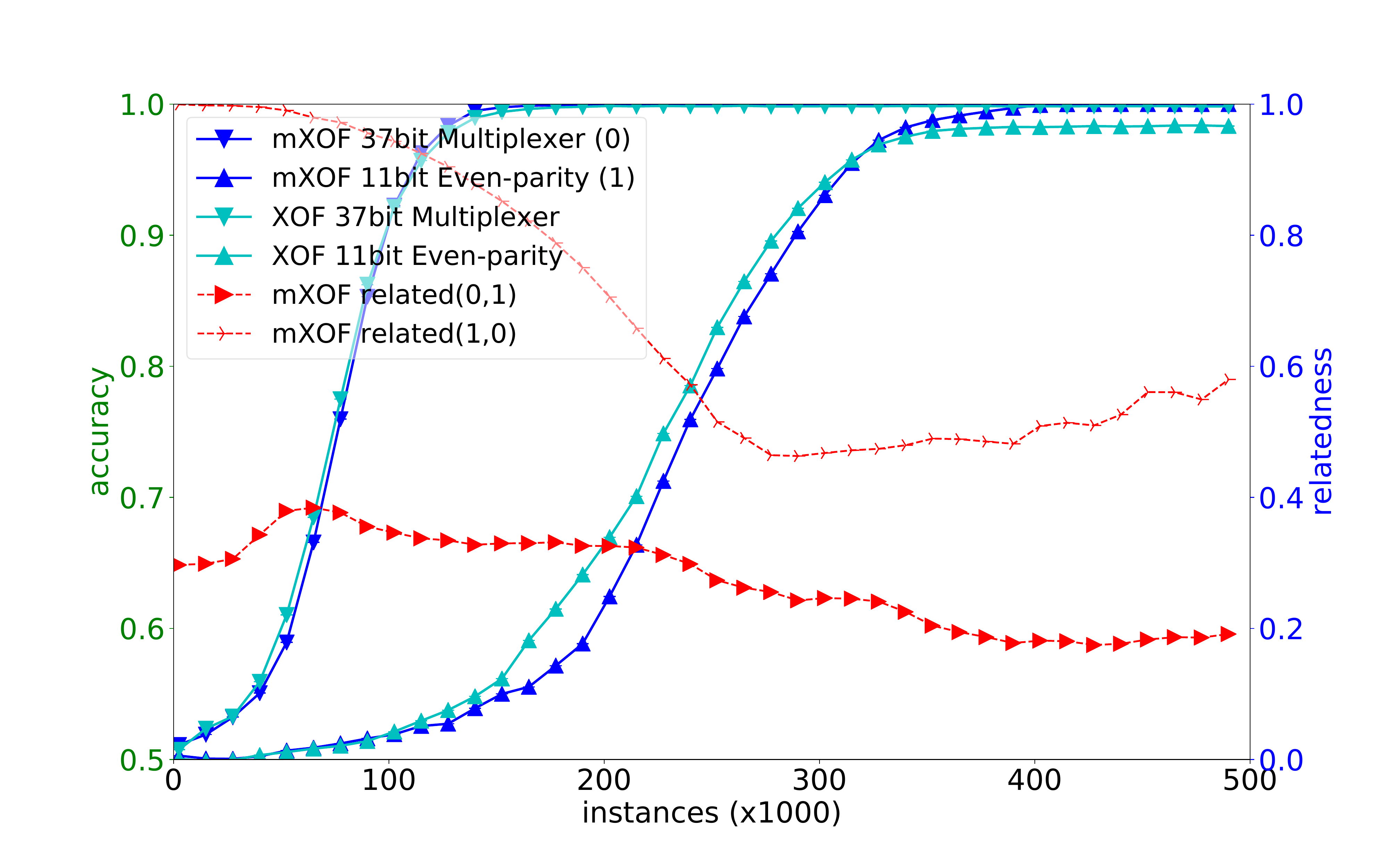}
	\caption{Learning performance on 37-bit Multiplexer and 11-bit Even-parity problems.}
	\label{fig:mux37_epar11}
	\vspace{-3mm}
\end{figure}

Figure \ref{fig:mux37_epar11} shows the learning performances of mXOF on 37-bit Multiplexer and 11-bit Even-parity problems together, and XOF on these two problems separately. The learning curves of mXOF and XOF on corresponding problems have no substantial difference. The learning process of mXOF on 11-bit Even-parity problem can even converge to $100\%$ faster. The relatedness of 37-bit Multiplexer task to 11-bit Even-parity starts low because their OLs start with base CFs encoding original data attributes. Specifically, the starting OL of 11-bit Even-parity system is $(D_0,...,D_{10})$, while $(D_0,...,D_{36})$ is the one in 37-bit Multiplexer system. Hence, only the part $(D_0,...,D_{10})$ in the OL of 37-bit Multiplexer system can be applicable to the other system. It then increases a little before declining together to the relatedness of around $0.2$. 

One of the $30$ runs of the single system on the 11-bit Even-parity problem was stuck at the accuracy of $50\%$. In this run, one classifier containing only one base CF in its condition dominated the rule population with very high numerosity and fitness. This was explained in \cite{nguyen2020complexity} as this problem poses a high chance of local optima for XOF because of the influence of CF-fitness in genetic operations. The diversity from external task, the 37-bit Multiplexer problem, helps XOF escape the local optima because solving the 37-bit Multiplexer problem only requires XOF to use base CFs, which cover all base CFs that the 11-bit Even-parity problem needs to balance before generalising.

\begin{figure}
	\vspace{-3mm}
\centering
\includegraphics[width=0.9\columnwidth]{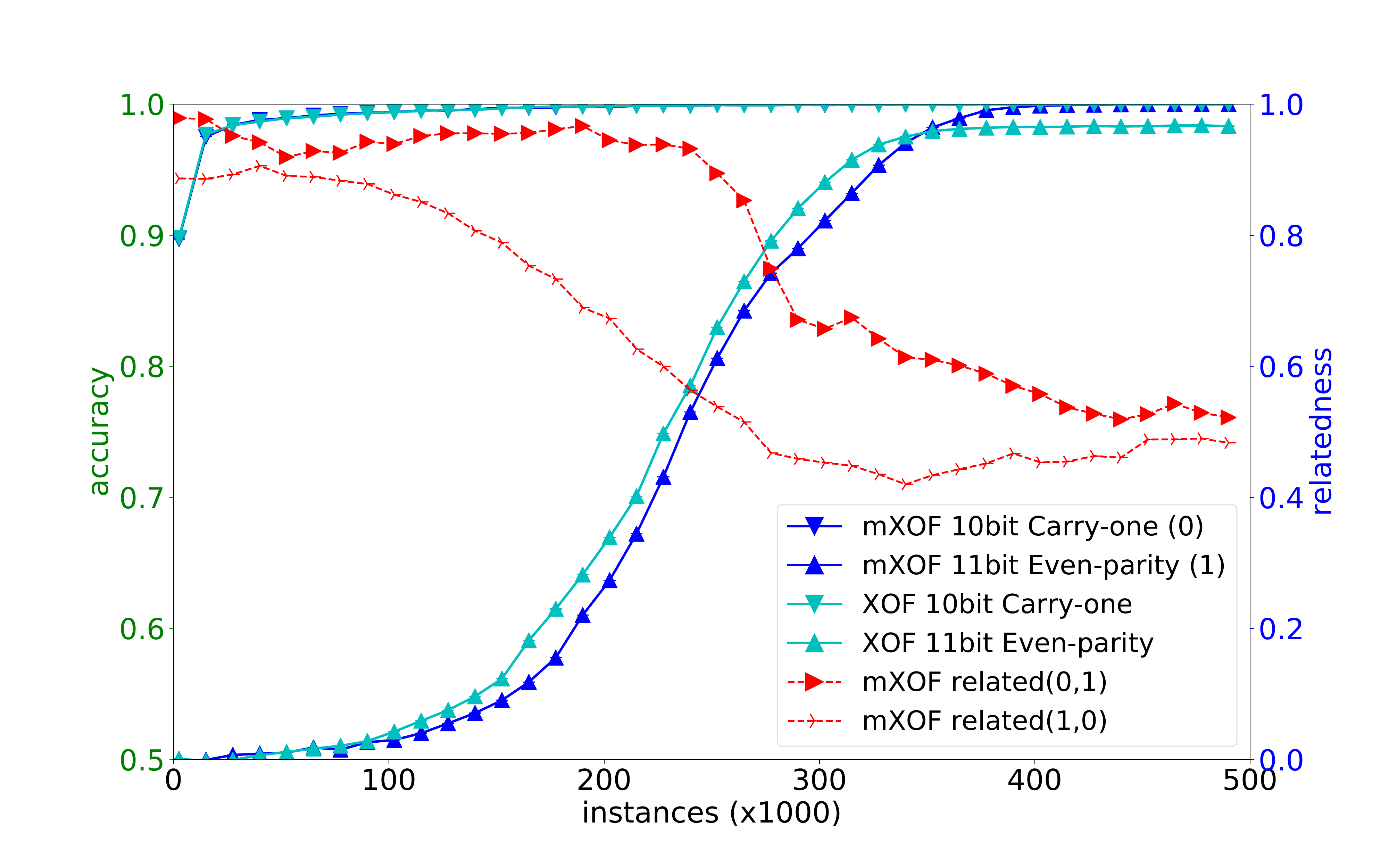}
\caption{Learning performance on 10-bit Carry-one and 11-bit Even-parity problems.}
\label{fig:car10_epar11}
\vspace{-3mm}
\end{figure}

Similar trends occur when learning multiple tasks with 10-bit Carry-one and 11-bit Even-parity problems (see Figure \ref{fig:car10_epar11}). The learning performances of both tasks in mXOF are no substantially different from learning them separately with XOF. Both relatedness on two directions stays relatively high in the beginning but then declines to the values of near $0.5$ when 11-bit Even-parity system starts progressing. The fact that these relatedness parameters stays high in the early phase of both cases in this section seems to suggest that this relatedness estimation does not reflect the relatedness of these tasks. We will discuss further in the Discussion.

\subsection{Multi-class Classification using Multiple Binary Classifiers}

In this section, we will provide an initial investigation on the performance of mXOF compared with several popular machine learning algorithms on the UCI Zoo dataset. All results were evaluated using 10-fold cross-validation in supervised learning. Most of the baseline algorithms were tested using Weka \cite{hall09:_weka_data_minin_softw} with default settings except for Multi-layer Perceptron (MLP) and Random Forest. Random Forest was set with the batch size of $200$, which was the best result among several tested batch sizes. MLP used two hidden layers of sizes $15$ and $10$. The one nominal attribute in this dataset is converted to multiple binary attributes using one-hot encoding. Except for Naive Bayes and C4.5 classifiers, which are not stochastic algorithms, we ran all other algorithms $30$ times with different random seeds to evaluate the averages and standard deviations of their accuracies. mXOF scores $95.83\%$ and $96.25\%$ on average with the population size of $N=500$ and $N=1000$ for each system respectively (see Table \ref{tab:uci_zoo}). These population sizes are considered relatively small for XCSs. These results are competitive compared with other popular machine learning algorithms in this experiment even though the differences are not statistically significant based on the Wilcoxon Signed-Ranks test with $p{-value}<0.05$.

\begin{table*}
	\vspace{-3mm}
	\centering
	\caption{Results on the UCI Zoo dataset in supervised learning.}
	\begin{tabular}{ |c|c|c|c|c|c|c|c| }
		\hline
		{\textbf{Problem}} & {\textbf{Naive Bayes}} & {\textbf{SVM}} & {\textbf{MLP}} & {\textbf{C4.5}} & {\textbf{Random Forests}} & {\textbf{mXOF} ($N=500\times 7$)} & {\textbf{mXOF} ($N=1000\times 7$)} \\ \hline
		zoo & $95.05\%$ & $92.08\%$ & $95.91 \pm 0.42\%$
		& $92.08\%$ & $96.07 \pm 0.65\%$ & $95.83 \pm 1.09\%$ & $96.25 \pm 1.30\%$ \\ \hline
	\end{tabular}
	\label{tab:uci_zoo}
	\vspace{-3mm}
\end{table*}

\begin{figure}
	\vspace{-1mm}
	\centering
	\includegraphics[width=1.0\columnwidth]{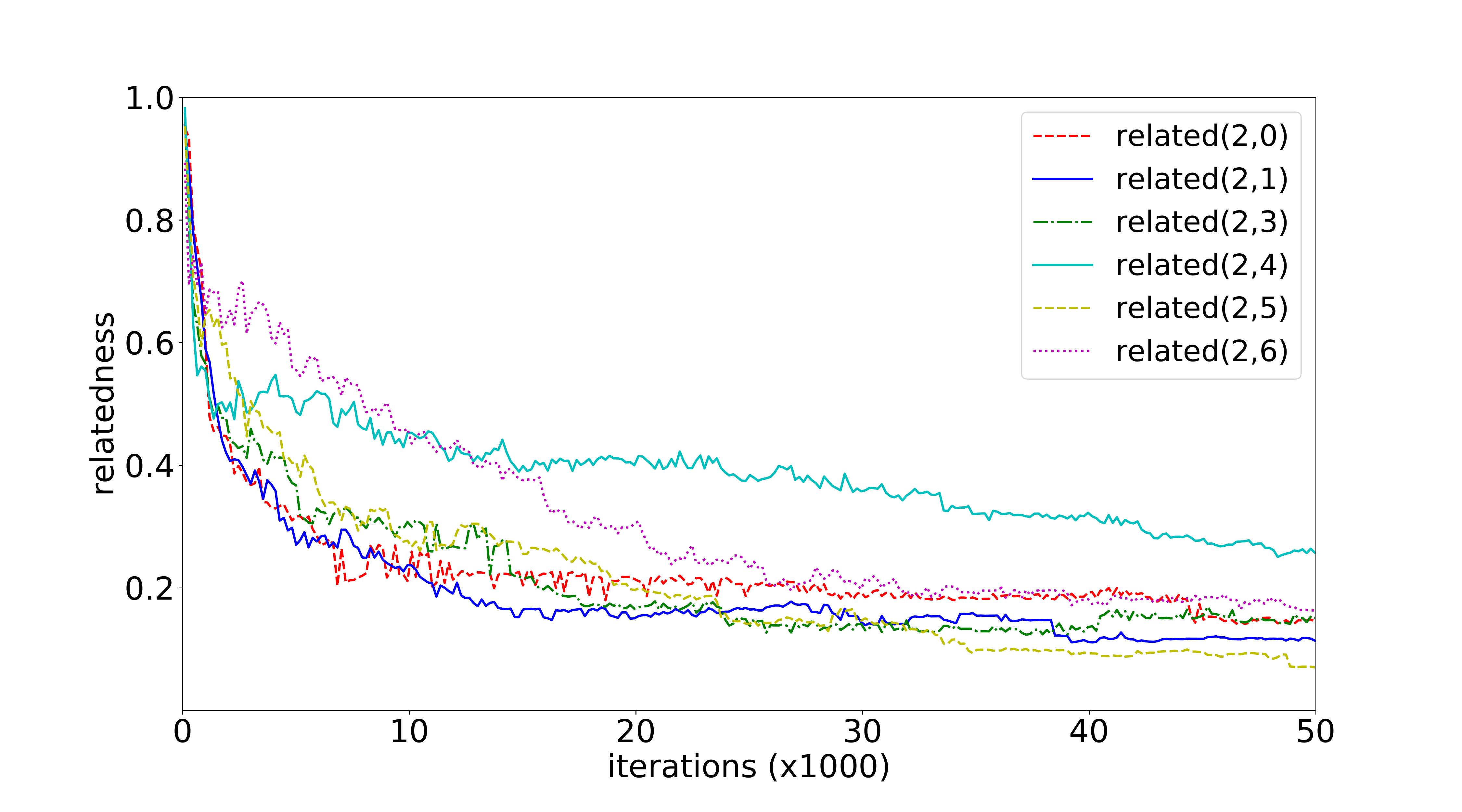}
	\caption{The dynamic flow of relatedness of task for class $2$ to other tasks (classes). Classes numbers correspond to actual classes as follows: $0$: mammal, $1$: bird, $2$: reptile, $3$: fish, $4$: amphibian, $5$: insect, and $6$: invertebrate.}
	\label{fig:uci_zoo_rel_flow}
	\vspace{-3mm}
\end{figure}
\begin{figure}
	\vspace{-1mm}
	\centering
	\includegraphics[width=0.9\columnwidth]{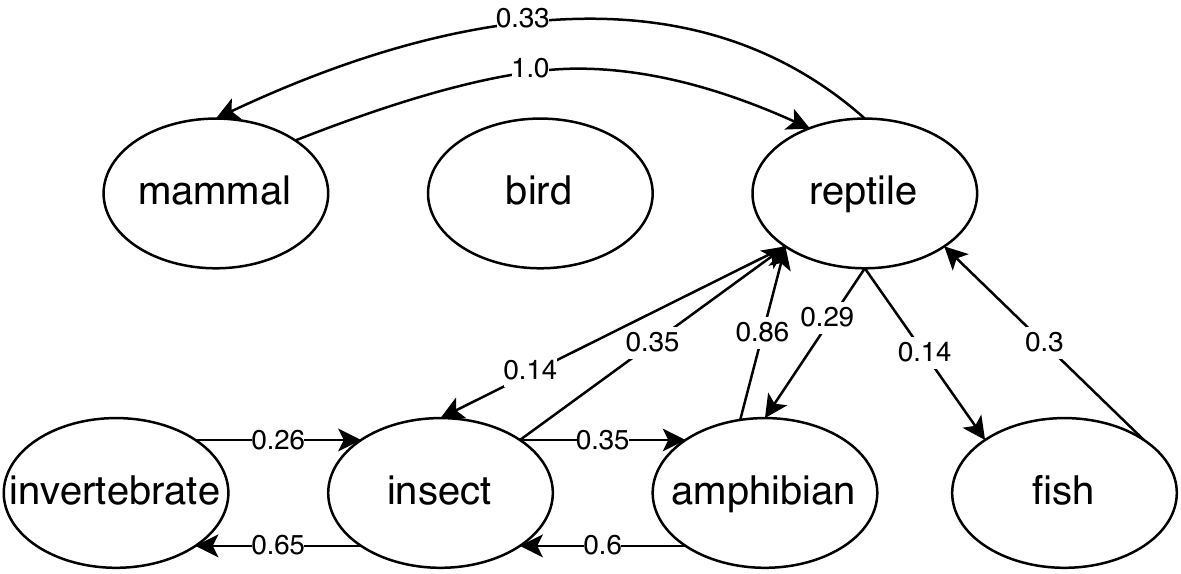}
	\caption{The final relatedness of $7$ classes in the UCI Zoo dataset. These are results in one of $30$ runs.}
	\label{fig:uci_zoo_rel}
	\vspace{-3mm}
\end{figure}

Figure \ref{fig:uci_zoo_rel_flow} illustrates the average relatedness of class ``reptile" to other classes. The initial class relatedness starts at near $1.0$ because all binary classifiers start with the OLs of all original data attributes. All the relatedness fell to the final optimal values equivalent to the final relatedness in Figure \ref{fig:uci_zoo_rel} as the systems generalise their rules with fewer discriminative features. The class ``reptile" has the most interactive relatedness with four other classes because the optimal rules for ``reptile" include the largest number of data attributes. Therefore, it has high chance to share common building blocks with other tasks. On the contrary, class ``bird" needs only attribute ``feathers" to be recognised from other classes. Also, this attribute is only needed for class ``bird". These two factors result in no link between ``bird" and other classes.

\subsection{Discussions}

The relatedness among tasks can guide the sharing of constructed knowledge among tasks in mXOF to improve the learning performances of related tasks. The learning processes of each task can benefit from useful CFs found in other tasks. The estimated relatedness among tasks can also reduce harmful interference among relatively unrelated tasks. As a result, the learning performances of these tasks remain unchanged compared with separate learning. On the other hand, the feature sharing among tasks also reinforces the relatedness when useful CFs become common among tasks and thereby dynamically change the relatedness parameters. 

The high values of relatedness in the early phase of the two experiments in Section \ref{ssect:low_related} do not show that these tasks are highly related. However, they do share common CFs in the early phase because, in this phase, each system has not grown complex CFs other than base CFs. This situation is similar to the beginning phase of human learning to recognise different objects using signals from the same sense. The decreases of these values happen quickly when the 11-bit Even-parity system starts building up complex patterns to replace original data attributes. This explains why two relatively unrelated tasks have such high relatedness in the early learning phase. Similarly, the relatedness of the 18-bit Hierarchical Carry-one task to the 18-bit Hierarchical Multiplexer task declines quickly to the values of around $0.6$ because these systems build up complex patterns with a limited divergence. They start to diverge with their own complex-pattern discovery including patterns specialised for their problems. The contradiction between the system divergence and feature sharing in mXOF causes the relatedness to balance at around $0.6$. In short, the relatedness parameter in our experiments describes reasonably the actual dynamics of constructing tree-based features.

The transferred CFs also provide each system in mXOF external diversity, which can be valuable in escaping local optima. The learning process of mXOF on 11-bit Even-parity problem when learning with 37-bit Multiplexer problem can even converge to $100\%$ faster than XOF because of the small external influence from the task solving 37-bit Multiplexer problem. In the case of highly related tasks (multiple hierarchical problems), this influence results in easier accuracy convergence of mXOF on 18-bit hierarchical problems.

mXOF can also work as a multi-class classifier with competitive results on the UCI Zoo dataset although it is slower due to being an online learning algorithm. The performance of mXOF tends to increase with larger population size. However, the optimal rules for this dataset mainly use $(AND,OR,NOT)$ logics with the original data attributes. The transferable building blocks are only the original attributes. Therefore, the benefits of growing and sharing complex features among tasks are not needed. However, the experiments of mXOF on this dataset are an initial investigation to show its potential in solving multi-class/multi-label classification and demonstrating the links among classes.

\section{Conclusions and Future Works}\label{sect:conclusions}
We have developed an MTL system using multiple XOFs with the ability to adapt the feature sharing among tasks automatically. By crafting internal parameters of the hypothesised task relatedness to guide the automatic feature transfer, mXOF can improve the learning performances of individual tasks when they are related, and reduce harmful signals from other tasks when they are not supportive to a target task. The relatedness parameter based on the OL of XOF estimate reasonably the dynamic commonality of patterns among tasks. The dynamic update of relatedness is essentially useful for learning systems with feature construction, e.g. mXOF, because the benefit of sharing features among tasks may only occur at some specific stages of feature-complexity growth. However, further development and investigation on mXOF is necessary to explore its abilities on a broader range of problems.

Having the problem relatedness measurements can help create network links among target objects of the binary classifiers in mXOF. Therefore, learning more objects builds up this knowledge network. This network enables links of only specific knowledge that could be useful for a target task. In an AI system with a high volume of accumulated knowledge, this ability is essential to avoid intractable search spaces when querying all knowledge.

Future research can consider mXOF for the context of continual and multitask learning. The reason is that in mXOF, learning a new class only requires spawning a new system without remarkable negative impacts on existing tasks given a proper estimation of relatedness. Learning a new class could take advantage of the bias of previously learned knowledge to acquire relevant knowledge within fewer examples. This is equivalent to human/robot learning to recognise multiple objects using signals from the same sense/sensor.

Because XCS and the OF module can be considered frameworks to be integrated with different representations (for its rules), mXOF is not bound to using only tree-based programs (CFs). Future research should consider integrating mXOF with neural networks to learn real-valued data. This combination could also be fruitful in producing arbitrary and complexity-efficient network structures.





\bibliographystyle{ACM-Reference-Format}
\bibliography{mXOF}

\end{document}